\documentclass{article}

\usepackage[preprint]{neurips_data_2024}

\usepackage[utf8]{inputenc} 
\usepackage[T1]{fontenc}    
\usepackage{hyperref}       
\usepackage{url}            
\usepackage{booktabs}       
\usepackage{amsfonts}       
\usepackage{nicefrac}       
\usepackage{microtype}      
\usepackage{xcolor}         
\usepackage{pifont}
\usepackage{multicol}
\usepackage{multirow}
\usepackage{colortbl}
\usepackage{array}
\usepackage{graphicx}
\usepackage{makecell}
\usepackage{caption}
\usepackage{subcaption}
\usepackage{longtable}
\usepackage{booktabs}
\usepackage{amsmath}

\title{Chem2Gen-Bench: Benchmarking Chemical-to-Genetic Translation in Perturbation Response Space}

\author{%
Yuxiang Lin\textsuperscript{1}
\qquad Ying Chen\textsuperscript{2}
\\
\textsuperscript{1}National Institute for Data Science in Health and Medicine, Xiamen University \\
\textsuperscript{2}School of Informatics, Xiamen University\\ 
{\tt\small linyuxiang@stu.xmu.edu.cn }
}

\begin{document}

\maketitle

\begin{abstract}
Virtual-cell and perturbation models are increasingly used to predict cellular responses for biomedical discovery, but chemical and genetic perturbations are not automatically interchangeable. Existing evaluations often study chemical response prediction or genetic perturbation prediction separately, leaving target-matched chemical-to-genetic translation under-tested. 
We introduce Chem2Gen-Bench, a benchmark comprising 260,084 chemical and 1,099,045 genetic perturbation profiles organized into cell-target contexts, and evaluate pairwise alignment, retrieval, protocol covariate associations, feature spaces, and foundation-model embeddings.
Across matched contexts, translation fidelity is measurable but heterogeneous; background adjustment increases the association between pairwise similarity and retrieval success, while paired tests show lower mean retrieval success after adjustment under the evaluated settings. 
In a target-matched K562 audit, the evaluated foundation-model embeddings did not consistently improve over gene-delta baselines. Chem2Gen-Bench provides an auditable framework for testing when chemical and genetic perturbations align around shared targets and when representation gains are supported by matched perturbation evidence.
\end{abstract}
\section{Introduction}

Virtual-cell modeling aims to build computational systems that predict cellular behavior under defined contexts and perturbations \cite{wei_vcworld_2025,dibaeinia_virtual_2026}. This goal is relevant to drug discovery because perturbation-response models can help prioritize compounds, targets, and experimental settings for follow-up \cite{srivatsan_massively_2020,qi_predicting_2024}. Perturbation response is a natural evaluation setting: interventions create measurable changes in molecular state, and those changes can be compared across contexts \cite{peidli_scperturb_2024}. The two most common intervention classes are chemical perturbations, which expose cells to compounds, and genetic perturbations, which alter target genes or their products \cite{mcfarland_multiplexed_2020,replogle_mapping_2022}.

Genetic perturbations are informative probes of target function, especially when pooled CRISPR screens are read out at single-cell resolution \cite{replogle_mapping_2022}. Chemical perturbations are closer to drug-discovery workflows, but compound responses depend on target engagement, dose, exposure time, off-target activity, and cell context \cite{srivatsan_massively_2020,mcfarland_multiplexed_2020}. The target therefore provides the bridge between modalities: if a compound is linked to a protein target, a genetic perturbation of the corresponding gene gives a reference response for target-level comparison \cite{corsello_drug_2017,chandrasekaran_three_2024}. This bridge is useful but incomplete, because matched chemical and genetic perturbations can remain difficult to align even in designed profiling resources \cite{chandrasekaran_three_2024}. The benchmark question is therefore how closely a chemical response matches the corresponding genetic response in the same cell-target context.

Existing resources and models support chemical response prediction, genetic perturbation prediction, and broad perturbation benchmarking \cite{qi_predicting_2024,roohani_predicting_2024}. These settings do not by themselves answer whether chemical and genetic perturbations linked by the same target align under matched cell-target contexts. Source effects, dose, time, cell context, and target specificity can create apparent agreement or disagreement, and Systema has shown that common perturbation-prediction metrics can be inflated by systematic variation \cite{vinas_torne_systema_2025}. Recent single-cell foundation models motivate comparisons in learned embedding spaces, but their benefit for target-matched chemical-to-genetic translation should be tested against simple baselines rather than assumed \cite{cui_scgpt_2024,theodoris_transfer_2023}.

Here we introduce Chem2Gen-Bench, a target-centered benchmark for chemical-to-genetic perturbation translation. The benchmark harmonizes 260,084 chemical and 1,099,045 genetic profiles into cell-target contexts and evaluates translation at matched, generalized, and specificity-oriented levels (Figure~\ref{fig:fig1}). Chem2Gen-Bench compares pairwise alignment, retrieval, null baselines, protocol covariate associations, target enrichment, and representation choices across gene, pathway, background-adjusted, and foundation-model embedding spaces. The results show that translation fidelity is context-specific, sometimes associated with observed protocol covariates, and not consistently improved by the evaluated foundation-model embeddings in the K562 target-matched audit. This framing treats chemical-to-genetic translation as a measurable benchmark problem rather than as a general claim about therapeutic substitutability.
\section{Related Work}

\subsection{Perturbation Atlases and Virtual Cells}

Perturbation atlases provide the empirical basis for virtual-cell evaluation because they measure how cells respond after controlled interventions. scPerturb harmonizes public single-cell perturbation-response datasets and supports method development across molecular readouts \cite{peidli_scperturb_2024}. The LINCS L1000 Connectivity Map provides a large bulk transcriptional perturbation resource for compounds and genetic reagents \cite{subramanian_next_2017}. Newer atlases increase scale and context diversity, including Tahoe-100M for single-cell chemical perturbations and broader efforts toward perturbation cell and tissue atlases \cite{zhang_tahoe-100m_2025,rood_toward_2024}. Virtual-cell position papers and models argue that predictive systems need context, perturbation coverage, and feedback from experimental data \cite{qian_grow_2025,dibaeinia_virtual_2026}. Chem2Gen-Bench uses this motivation but asks a narrower question: whether matched chemical and genetic responses align around the same target.

\subsection{Chemical Perturbation Profiling and Target Resources}

Chemical perturbation profiling measures compound-induced molecular or cellular-state changes. Massively multiplex chemical transcriptomics showed that single-cell readouts can capture heterogeneous compound responses at scale \cite{srivatsan_massively_2020}. Multiplexed cancer-cell response profiling further showed that drug responses vary across cell contexts and can inform mechanism-of-action analysis \cite{mcfarland_multiplexed_2020}. Drug-target resources such as the Drug Repurposing Hub help connect compounds to annotated targets, but target annotations remain incomplete and do not remove off-target biology \cite{corsello_drug_2017}. Recent chemical-response models and benchmarks extend this direction with generative prediction, dynamic drug-response modeling, and drug-response foundation-model audits \cite{qi_predicting_2024,guo_scstatedynamics_2024,wang_scdrugmap_2025}. Deep generative approaches for disease-state dynamics also motivate in silico drug-discovery analyses, although they do not directly test target-matched chemical-to-genetic translation \cite{zheng_deep_2025}. Chem2Gen-Bench differs by using the target link to compare compound responses against genetic perturbation responses rather than predicting chemical response alone.

\subsection{Genetic Perturbation Prediction and Representation Learning}

Genome-scale Perturb-seq makes genetic perturbation response measurable at single-cell resolution \cite{replogle_mapping_2022}. This has enabled models for genetic perturbation prediction, including graph-enhanced response prediction, neural optimal transport, variational response modeling, and disentangled latent representations \cite{roohani_predicting_2024,bunne_learning_2023,lotfollahi_scgen_2019,piran_disentanglement_2024}. Causal and statistical analyses of single-cell perturbation atlases further emphasize that measured perturbation effects can depend on confounding structure, power, and effect-size heterogeneity \cite{dong_causal_2023,nadig_transcriptome-wide_2025}. Other methods model complex or combinatorial perturbations and heterogeneous response structure \cite{lotfollahi_predicting_2023,song_decoding_2025}. Single-cell foundation models such as scGPT and Geneformer motivate embedding-space evaluation, and perturbation-trained systems such as State and Tahoe-x1 increase the scale of learned representations \cite{cui_scgpt_2024,theodoris_transfer_2023,adduri_predicting_2025,gandhi_tahoe-x1_2025}. Chem2Gen-Bench treats these embeddings as candidate representations and tests them against gene-delta baselines in the available target-matched K562 audit.

\subsection{Benchmarking Chemical-Genetic Correspondence}

Matched chemical-genetic benchmarks have been most developed in image-based profiling. CPJUMP1 and the broader JUMP Cell Painting resource were designed to evaluate perturbation similarity across chemical and genetic interventions, and these studies report that matching perturbations around shared targets remains challenging \cite{chandrasekaran_three_2024,chandrasekaran_jump_2023}. Later Cell Painting resources expanded genetic coverage and profiling quality for morphology-based perturbation analysis \cite{seal_cell_2025,chandrasekaran_morphological_2025}. Generative morphology models provide another route for perturbation-response prediction in imaging spaces \cite{palma_predicting_2025}. In transcriptomic perturbation prediction, Systema, PerturBench, and related benchmarking studies emphasize systematic variation, simple baselines, and metric sensitivity \cite{vinas_torne_systema_2025,wu_perturbench_2025,li_benchmarking_2025}. Independent comparisons also show that new perturbation models must be judged against transparent baselines under matched settings \cite{ahlmann-eltze_deep-learning-based_2025,li_systematic_2024,wei_benchmarking_2025}. These works motivate Chem2Gen-Bench: a transcriptomic, target-matched, protocol-aware benchmark for chemical-to-genetic translation.

\begin{figure*}[!t]
\centering
\includegraphics[width=\textwidth]{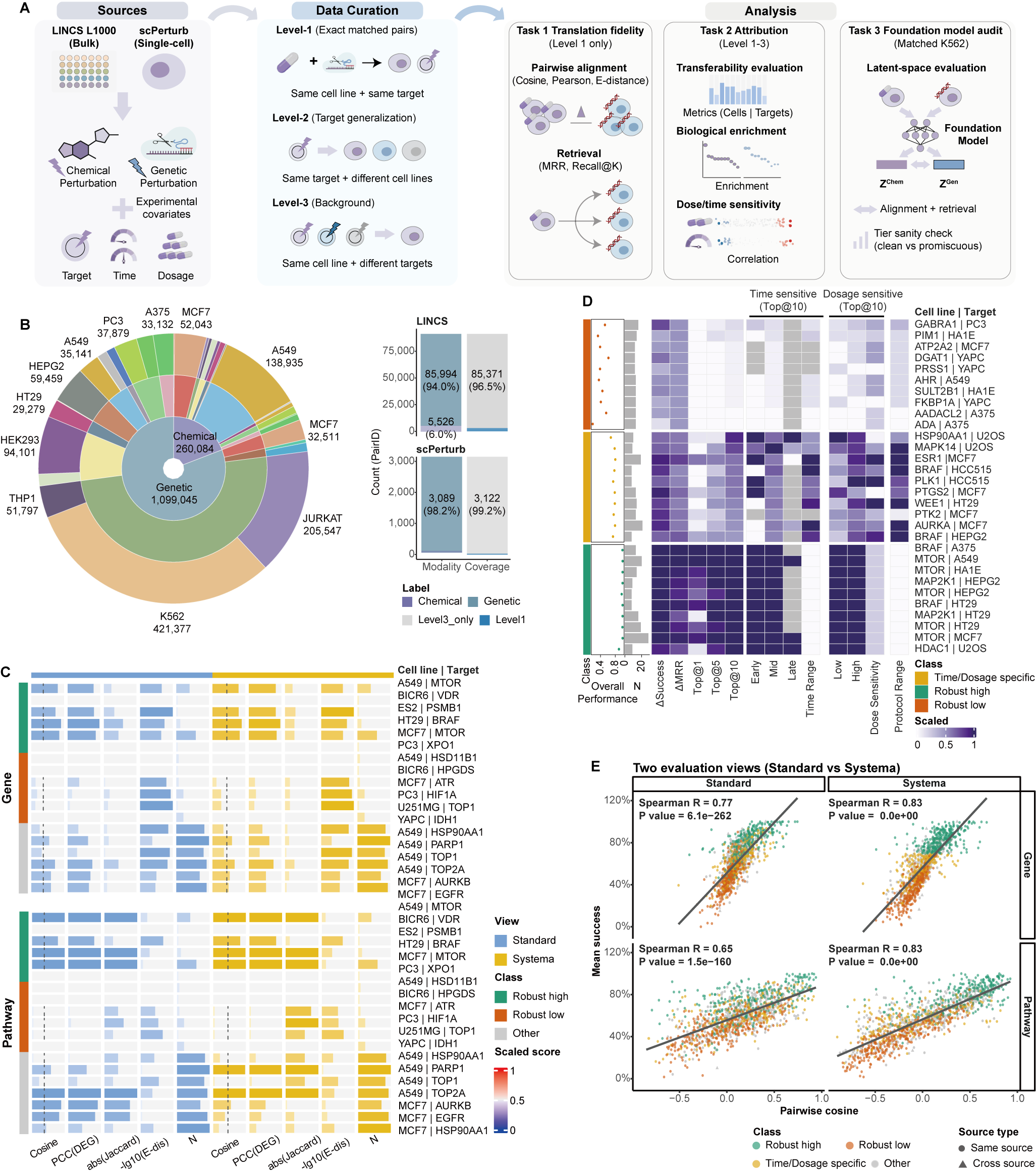}
\caption{\textbf{Chem2Gen-Bench construction and benchmark design.} (A) Workflow from perturbation profiles to representation construction, cell-target contexts, benchmark levels, pairwise alignment, retrieval, robustness classes, and representation audit. (B) Resource composition, including 260,084 chemical perturbation profiles and 1,099,045 genetic perturbation profiles. (C) Pairwise alignment across matched contexts. (D) Retrieval fingerprints and descriptive robustness classes. (E) Relationship between pairwise cosine and retrieval success under Standard and Systema-style views.}
\label{fig:fig1}
\end{figure*}
\section{Methods}

\subsection{Benchmark Organization}

Chem2Gen-Bench organizes chemical and genetic perturbation profiles into target-centered evaluation units. Each perturbation instance \(i\) has a source or study label \(s\), a cell context \(c\), a target gene \(g\), a perturbation modality \(m \in \{\mathrm{chem}, \mathrm{gen}\}\), and an observed profile or embedding vector \(x_i\). A context is defined as \(k=(c,g)\). Level 1 evaluates exact chemical-to-genetic translation within the same cell-target context. Level 2 evaluates target-linked generalization across cell contexts. Level 3 evaluates specificity against alternative targets in the same cell context.

\subsection{Perturbation-Effect Representations}

All evaluations operate on perturbation-effect vectors relative to matched controls. For instance \(i\), the base delta is
\[
\Delta_i = x_i - \bar{x}_{0,s_i,c_i},
\]
where \(x_i\) is the observed expression or embedding vector and \(\bar{x}_{0,s_i,c_i}\) is the mean control vector matched to source \(s_i\) and cell context \(c_i\). Gene-space deltas use \(\Delta_i\) directly. Pathway-space deltas are obtained as
\[
z^{\mathrm{path}}_i = W^\top \Delta^{\mathrm{gene}}_i,
\]
where \(W\) is a gene-by-pathway membership or weighting matrix after matching genes to the expression feature space. For background-adjusted analyses, the selected representation \(z_i\) is centered against a matched background stratum following the logic of systematic-variation-aware evaluation \cite{vinas_torne_systema_2025}:
\[
\tilde{z}_i = z_i -
\frac{1}{|\mathcal{B}(b_i)|}\sum_{j \in \mathcal{B}(b_i)} z_j .
\]
Here \(\mathcal{B}(b_i)\) is the set of background instances in stratum \(b_i\). In the current result files, background summaries are indexed by representation track, source database, perturbation modality, and standardized cell line; final submission should verify whether this exact stratum or a stricter target-exclusion variant was used for every Systema-style panel.

\subsection{Alignment and Distribution Distances}

For each context and modality, Chem2Gen-Bench computes a centroid in the selected representation space:
\[
\bar{z}_{k,m} =
\frac{1}{N_{k,m}}\sum_{i \in \mathcal{I}_{k,m}} z_i,
\]
where \(k=(c,g)\), \(m \in \{\mathrm{chem}, \mathrm{gen}\}\), \(\mathcal{I}_{k,m}\) is the set of instances in that context and modality, and \(N_{k,m}=|\mathcal{I}_{k,m}|\). Pairwise alignment compares \(\bar{z}_{k,\mathrm{chem}}\) with \(\bar{z}_{k,\mathrm{gen}}\):
\[
\cos_k =
\frac{\bar{z}_{k,\mathrm{chem}}^\top \bar{z}_{k,\mathrm{gen}}}
{\|\bar{z}_{k,\mathrm{chem}}\|_2 \|\bar{z}_{k,\mathrm{gen}}\|_2}.
\]
When instance distributions are compared directly, energy distance is used because it measures distances between two sample sets and has been used for single-cell perturbation-effect comparison \cite{szekely_energy_2013,peidli_scperturb_2024}. For chemical set \(X=\{x_i\}_{i=1}^{n}\) and genetic set \(Y=\{y_j\}_{j=1}^{m}\),
\[
\begin{aligned}
E(X,Y) ={}&
\frac{2}{nm}\sum_{i=1}^{n}\sum_{j=1}^{m}\|x_i-y_j\|_2 \\
&- \frac{1}{n^2}\sum_{i=1}^{n}\sum_{j=1}^{n}\|x_i-x_j\|_2 \\
&- \frac{1}{m^2}\sum_{i=1}^{m}\sum_{j=1}^{m}\|y_i-y_j\|_2 .
\end{aligned}
\]
Lower energy distance indicates closer empirical distributions in the selected representation.

\subsection{Retrieval and Null Calibration}

For a query with gallery size \(L\), tied ranks are averaged:
\[
\begin{aligned}
r_q &= \frac{r_{\min}(q) + r_{\max}(q)}{2}, \\
\mathrm{MRR}_q &= \frac{1}{r_q}, \quad
\mathrm{Success}_q = 1 - \frac{r_q-1}{L-1}.
\end{aligned}
\]
Here \(r_{\min}(q)\) and \(r_{\max}(q)\) are the first and last tied ranks of the correct gallery item for query \(q\). Null baselines are computed by permuting label identities while keeping similarity scores fixed. The null calibration table reports mean null MRR and Success for each track and view, and the Results state when displayed values are raw retrieval metrics rather than chance-corrected scores.

\subsection{Robustness, Protocol, and Enrichment Analyses}

\begin{figure*}[!t]
\centering
\includegraphics[width=\textwidth]{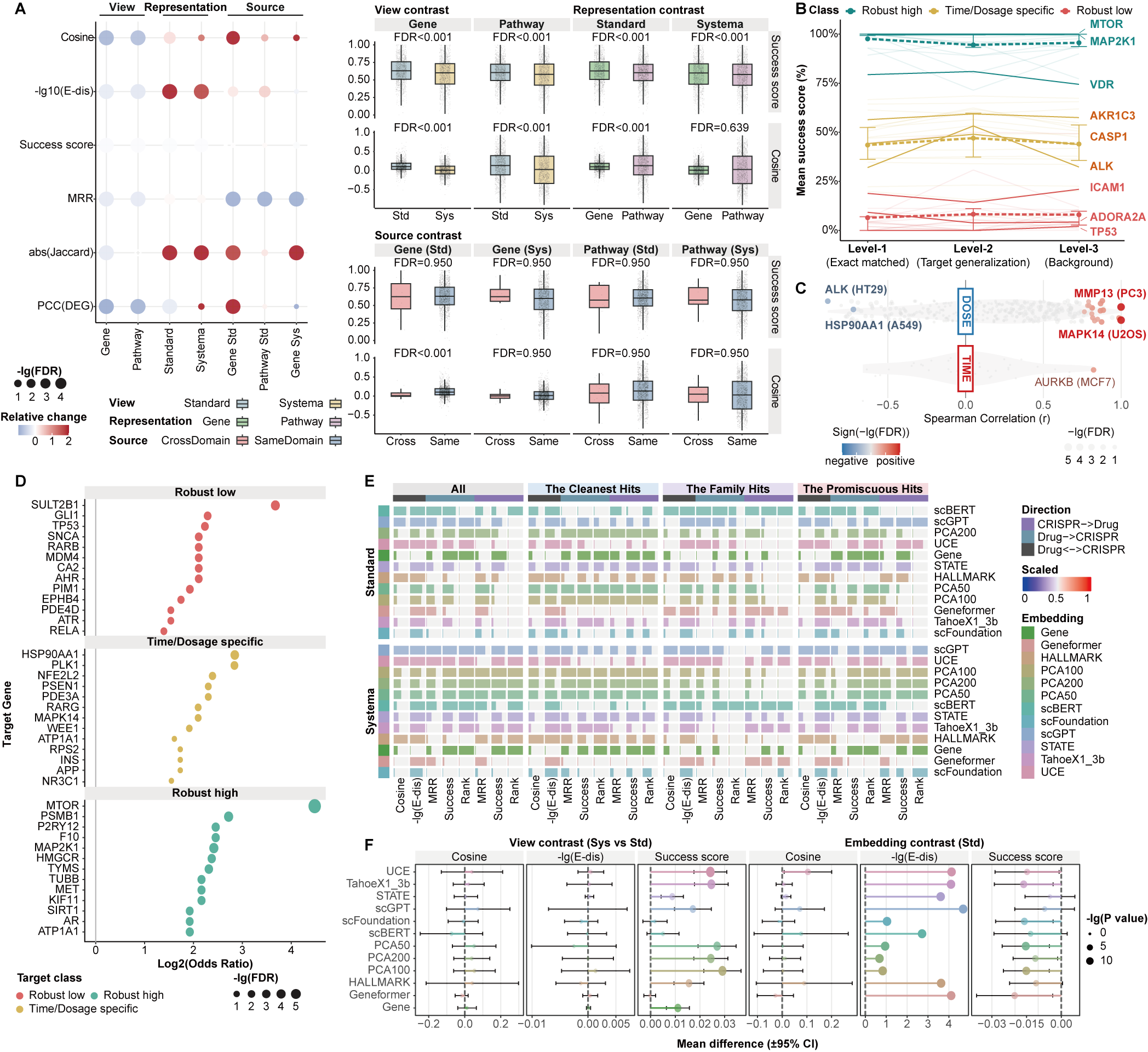}
\caption{\textbf{Representation contrasts, protocol associations, and embedding audit.} (A) Paired representation contrasts for Standard versus Systema-style views and gene versus pathway feature spaces. (B) Descriptive class trajectories across benchmark levels. (C) Dose and time associations with retrieval behavior in selected contexts. (D) Robust-high target enrichment. (E) K562 foundation-model audit leaderboard by target-specificity tier. (F) Baseline-relative differences with intervals derived across data folds for embedding and view contrasts.}
\label{fig:fig2}
\end{figure*}

Robustness classes are post hoc descriptive summaries of observed retrieval behavior. The current pipeline bins context-level \texttt{Mean\_Success} and \texttt{Peak\_Success} into tertiles. Contexts with high mean and high peak tiers are labeled \texttt{Robust\_High}; contexts with low peak tiers or low-low behavior are labeled \texttt{Robust\_Low}; contexts with low mean but high peak behavior are labeled \texttt{Protocol\_Sensitive}; the remaining combinations are labeled Intermediate. Target enrichment tests whether a target is overrepresented in a class using one-sided Fisher exact tests with Haldane-Anscombe corrected log odds ratios and Benjamini-Hochberg correction. Protocol analyses treat dose as ordered numeric exposure and time as ordered exposure duration within context, with Spearman correlations FDR-corrected across the tested protocol-sensitive context family. The foundation-model audit uses the available K562 target-matched chemical and CRISPR setup. The evaluated representation conditions include gene deltas, pathway summaries, principal-component baselines, and pretrained or perturbation-trained single-cell embedding spaces. Scores are computed for matched retrieval directions, metrics, and target-specificity tiers, and Figure~\ref{fig:fig2}F summarizes baseline-relative differences with intervals derived across data folds. For each embedding condition \(e\), baseline-relative change is summarized as
\[
\Delta_e = \mathrm{Score}_e - \mathrm{Score}_{\mathrm{baseline}},
\]
where the baseline is the matched gene-delta representation for the same retrieval direction, metric, and target-specificity tier. Paired Wilcoxon signed-rank tests are used for matched context comparisons, Mann-Whitney U tests for unpaired same-source versus cross-source comparisons, and Spearman correlations for dose and time associations. Benjamini-Hochberg false discovery rate correction is applied within each multiple-testing family \cite{benjamini_controlling_1995}.

\section{Results}

\subsection{Benchmark Scope and Evaluation Units}

Chem2Gen-Bench converts heterogeneous chemical and genetic perturbation profiles into comparable perturbation-effect instances. Each evaluation unit is organized by a cell-target context, and the benchmark separates exact matched translation, target-linked generalization, and specificity-oriented comparisons through the Level 1, Level 2, and Level 3 settings. The resource contains 260,084 chemical perturbation profiles and 1,099,045 genetic perturbation profiles, as shown in Figure~\ref{fig:fig1}. These definitions make chemical-to-genetic translation testable at exact, generalized, and specificity-oriented levels; they do not by themselves imply that any perturbation pair is biologically interchangeable.

\subsection{Translation Fidelity Is Heterogeneous}

We first evaluated whether chemical and genetic perturbations aligned uniformly in exact cell-target contexts. Pairwise alignment metrics in Figure~\ref{fig:fig1}C and retrieval fingerprints in Figure~\ref{fig:fig1}D show substantial variation across targets and cell lines. Some contexts show higher pairwise concordance and retrieval success, whereas others remain weak under the same benchmark framework. This pattern supports a restricted conclusion: translation fidelity is context-specific under the evaluated transcriptomic representations.

For exact Level 1 Chem-to-Gene target retrieval, raw observed retrieval metrics were compared with label-permutation null summaries. In the gene Standard view, observed mean success was 0.563 and observed mean MRR was 0.0677 across 68,541 queries, compared with null means of 0.498 and 0.0236. In the pathway Standard view, observed mean success was 0.536 and observed mean MRR was 0.0415, compared with null means of 0.495 and 0.0227.

\subsection{Representation Choices Change Metric Behavior}

Figure~\ref{fig:fig2} summarizes the representation, protocol, enrichment, and foundation-model analyses that follow the primary benchmark construction in Figure~\ref{fig:fig1}. Panel A reports paired representation contrasts, panels B-D summarize target-level and protocol-associated heterogeneity, and panels E-F audit learned embeddings in the K562 target-matched setting.

We compared Standard and Systema-style background-adjusted representations. In Figure~\ref{fig:fig1}E, the Spearman association between pairwise cosine and retrieval success increased from 0.77 to 0.83 in gene space and from 0.65 to 0.83 in pathway space after background adjustment. However, paired retrieval tests in Table~\ref{tab:contrasts} show lower mean success after adjustment in the signed Systema-minus-Standard contrast. The median difference was -0.01605 for gene mean success (\(N=2243\), FDR \(=4.25\times10^{-10}\)) and -0.01991 for pathway mean success (\(N=2243\), FDR \(=7.33\times10^{-10}\)). Thus, background adjustment increased consistency between pairwise alignment and retrieval utility in Figure~\ref{fig:fig1}E, but it reduced mean retrieval success in these paired comparisons.

\begin{table}[t]
\small
\centering
\caption{\textbf{Main paired representation contrasts.} Median difference uses the signed comparison stated in the first column. Each contrast used \(N=2243\) matched paired contexts.}
\label{tab:contrasts}
\begin{tabular}{@{}lrr@{}}
\toprule
Comparison & Median diff. & FDR \\
\midrule
Systema-Standard, gene & -0.01605 & \(4.25\times10^{-10}\) \\
Systema-Standard, pathway & -0.01991 & \(7.33\times10^{-10}\) \\
Pathway-Gene, Standard & -0.00617 & 0.00181 \\
Pathway-Gene, Systema & -0.01120 & \(3.51\times10^{-6}\) \\
\bottomrule
\end{tabular}
\end{table}

We next compared gene and pathway representations under Standard and Systema views. Table~\ref{tab:contrasts} shows that pathway-minus-gene mean success was negative under both settings: -0.00617 under Standard and -0.01120 under Systema. These paired results indicate lower target retrieval success for the pathway representation in the tested mean-success metric, but they do not imply that pathway aggregation is worse for every alignment statistic. Same-domain versus cross-domain comparisons were not detectably different after correction, but the cross-domain support was small (\(N_{\mathrm{cross}}=16\) versus \(N_{\mathrm{same}}=2227\)). We therefore treat the source-domain comparison as inconclusive.

\subsection{Robustness, Protocol, and Target Signals}

To summarize heterogeneous retrieval behavior, we assigned descriptive robustness classes based on observed benchmark patterns. Figure~\ref{fig:fig1}D displays the retrieval fingerprint and class labels, and Figure~\ref{fig:fig2}B tracks class behavior across benchmark levels. Robust-high, robust-low, and protocol-covariate-associated groups show distinct retrieval profiles, indicating that a single global average would hide important target-level structure.

We analyzed dose and time as observed protocol covariates within contexts. Figure~\ref{fig:fig2}C identifies selected contexts where ordered protocol variables were associated with retrieval behavior. A375-EGFR showed a positive dose association (\(\rho=0.868\), FDR \(=0.0070\)), HT29-WEE1 showed a positive dose association (\(\rho=0.884\), FDR \(=0.0068\)), and MCF7-AURKB showed a positive time association (\(\rho=0.822\), FDR \(=0.00957\)). These examples support the narrower claim that protocol covariates are associated with retrieval success in selected contexts, not that dose or time causally explains translation fidelity across the benchmark.

We tested whether selected targets were overrepresented among robust-high contexts. Figure~\ref{fig:fig2}D shows three robust-high examples that pass FDR correction. MTOR had 14 positives among 16 eligible contexts, with a robust-high rate of 0.875, \(\log_2\mathrm{OR}=4.48\), and FDR \(=2.27\times10^{-6}\). PSMB1 had 9 positives among 14 contexts, with a rate of 0.643, \(\log_2\mathrm{OR}=2.72\), and FDR \(=0.0241\). MAP2K1 had 10 positives among 17 contexts, with a rate of 0.588, \(\log_2\mathrm{OR}=2.42\), and FDR \(=0.0241\). These enrichments are associative summaries of robust-high behavior and should not be read as mechanistic explanations.

\subsection{Foundation-Model Embeddings in K562}

Finally, we audited learned representations in the available K562 target-matched chemical and CRISPR setup. Figure~\ref{fig:fig2}E ranks embedding conditions by target-specificity tier, and Figure~\ref{fig:fig2}F reports baseline-relative differences with intervals derived across data folds. Performance varies by model, metric, direction, and target-specificity tier. In the Standard CRISPR-to-Drug normalized-success contrast, seven evaluated learned/model tracks had negative mean differences versus the gene baseline across 2,372 queries, ranging from -0.0202 for Geneformer to -0.00491 for STATE. Under this K562 audit, the evaluated foundation-model embeddings did not consistently improve upon gene-delta baselines. This conclusion supports baseline-relative evaluation, not a broad dismissal of foundation models for other perturbation tasks.
\section{Discussion}

Chem2Gen-Bench evaluates target-matched chemical-to-genetic translation in perturbation-response space. The results indicate that translation is measurable but heterogeneous across cell-target contexts. Representation choices and foundation-model embeddings change the observed metric structure, but they do not remove the need for simple baselines and matched evaluation. Thus, chemical-to-genetic translation should be treated as a context-dependent benchmark problem.

The target bridge is central because it connects mechanistic genetic probes with compound exposures \cite{chandrasekaran_three_2024,peidli_scperturb_2024,replogle_mapping_2022}. However, target identity alone does not determine the observed response. A compound response also reflects target engagement, dose, time, off-target effects, and cell context \cite{srivatsan_massively_2020,mcfarland_multiplexed_2020,qi_predicting_2024}. Chem2Gen-Bench therefore evaluates transcriptomic correspondence around shared targets, not substitutability between genetic perturbation and compound treatment.

Protocol and representation patterns help illustrate why a single global ranking is insufficient. Dose and time associations in selected contexts are compatible with protocol-related variation, but they are observational associations rather than causal estimates. Background adjustment and pathway aggregation changed metric behavior, which supports reporting results by metric and representation instead of summarizing the benchmark as one winner. These constraints are consistent with prior cautions about systematic variation and baseline sensitivity in perturbation evaluation \cite{vinas_torne_systema_2025,ahlmann-eltze_deep-learning-based_2025,wu_perturbench_2025,li_benchmarking_2025}.

Single-cell foundation models provide a reason to compare learned representations for perturbation tasks, especially as large perturbation-trained models become more available \cite{cui_scgpt_2024,theodoris_transfer_2023,adduri_predicting_2025,gandhi_tahoe-x1_2025}. Chem2Gen-Bench tests a narrower question: whether embeddings improve target-matched chemical-to-genetic retrieval over gene-delta baselines. In the available K562 audit, gains varied by model, metric, direction, and target-specificity tier, and the evaluated embeddings did not consistently improve over gene-delta baselines. This result supports baseline-relative evaluation, not a general dismissal of foundation models.

Several limitations bound the interpretation. The benchmark combines measurement regimes that are not identical, including bulk and single-cell profiles. Drug-target annotations are incomplete and may miss off-target biology. Exact cell-target matching improves interpretability but reduces coverage. Protocol covariates are observed rather than randomized, so dose and time analyses should not be read as causal estimates. The foundation-model audit is restricted to the available K562 target-matched setting and should not be generalized to all cell types, perturbation tasks, or models. Chem2Gen-Bench provides an auditable framework for testing chemical-to-genetic translation and representation gains under matched perturbation settings.

\bibliographystyle{abbrvnat}
\bibliography{main}

\end{document}